# Transformer Based Implementation for Automatic Book Summarization

**Siddhant Porwal[1], Laxmi Bewoor*[2], Vivek Deshpande[3]**

*Abstract:* Document Summarization is the procedure of generating a meaningful and concise summary of a given document with the inclusion of relevant and topic-important points. There are two approaches- one is picking up the most relevant statements from the document itself and adding it to the Summary known as Extractive and the other is generating sentences for the Summary known as Abstractive Summarization. Training a machine learning model to perform tasks that are time-consuming or very difficult for humans to evaluate is major challenge. Book summarization is one of the complex tasks which is time consuming as well. Traditional machine learning models are getting modified with pre-trained transformers. Transformer based Language models trained in a self-supervised fashion gaining a lot of attention when fine-tuned for Natural Language Processing(NLP) downstream task like text summarization. This work is an attempt to use Transformer based technique for Book Summarization.

*Keywords: Summarization, Extractive, Abstractive, Transformer*

## 1. INTRODUCTION

The internet is now accessible through a variety of gadgets, putting it within reach of the general public. Since so much data on the internet is in an unstructured manner, it might be difficult to select exactly the information that is relevant and needed. Information overhead is a very common problem during this searching through a vast pool of information and normally this is a time consuming activity. However automatic text summarization is providing a helpful hand for getting the main notion within a shorter time-span [1].

Text Summarization is predominantly classified as extractive and abstractive summarization. Extractive summarization generates the summary based on the original text whereas Abstractive summarization constructs the summary by understanding and paraphrasing the text and proven to be a major challenge [2]. Deep Learning based models seem to be promising and providing hope to solve abstractive summarization challenges [3]. Though deep learning models like Recurrent Neural Network(RNN), Long Short Term Memory (LSTM) gained popularity for Natural Language Processing (NLP) tasks yet having the limitation of vanishing gradient and slower training. The development of Transformer based architecture proved to be a breakthrough that has led to the evolution of Text Summarization techniques [4-5]. The launch of Transformer and its subsequent improvements led to models like Bidirectional Encoder Representation from Transformers(BERT) which cause a meteoric rise of Transformer for NLP models [6].

This paper presents BERT based abstractive text summarization technique for book summarization. The paper is organized as follows: Section 2 describes related work for abstractive summarization. Section 3 describes BERT architecture .Section 4 describes BERT models. Section 5 elaborates BERT implementation and evaluation metric for book summarization.
Section 6 concludes the paper.

## 2. RELATED WORK

Abstractive summarization techniques are challenging as the summary uses new phrases or words reflecting the gist of the entire text very similar to human beings. Book summarization is yet another critical, time consuming and creative task which needs a contextually rich and relevant abstract which can attract readers to read the entire book. Despite this fact very little contribution has been witnessed for automatic book summarization [7-8]. This section is providing a brief review of work in abstractive summarization however no significant contribution is witnessed for the book summarization.

Peter Luhn[9], working at IBM in 1958 introduced the first text summarization algorithm. Luhn's algorithm was a naive approach based on Term Frequency (TF)- Inverse Document Frequency (IDF). The main concern was to look at the "window size" of non-important and highly important words. The maximum weight was assigned to sentences occurring near the starting of a document. Further, Akshil Kumar et al. [10] evaluated and the performance of three different algorithms were compared. The performance of three different algorithms was analyzed and compared by the author. First, the various text summarization techniques were explained. To extract important keywords for inclusion in the summary, extraction-based techniques were used. Three keyword extraction algorithms were used for comparison: TextRank, LexRank, and Latent Semantic Analysis (LSA). The evaluation metric Recall-Oriented Understudy for Gisting Evaluation (ROUGE)-1 was applied to assess the efficacy of the extracted keywords. The algorithms' output was compared to the handwritten-summaries to assess performance and it was concluded that the TextRank algorithm outperformed the other two. The solutions for automated text summarization were attempted with statistical and machine learning based models based on factors like position in the documents, similarity with the title, TF*IDF etc.[11]. Andhale and Bewoor[1] reviewed various other approaches to deal with text summarization. This section is mainly highlighting deep learning models used for abstractive summarization.

[1] *Vishwakarma Institute of Information Technology, Pune-411048,INDIA*
*ORCID ID : 0000-0002-8918-2466*
[2] *Vishwakarma Institute of Information Technology, Pune-411048,INDIA*
*ORCID ID : 0000-0003-3470-421X*
[3] *Vishwakarma Institute of Information Technology, Pune-411048,INDIA*
*ORCID ID : 0000-3343-7165-777X*
*\* Corresponding Author Email: laxmi.bewoor@viit.ac.in*

The major focus of deep-learning is to extract features in data driven perspective rather than manual feature identification. The depth of neural networks helps to classify the importance of statements for the summary generation. Syed [12] discussed the use of RNN, deep learning model for processing the data in sequential order. The embedding of words, sentences, phrases were the input for the RNN and the output was the word embedding of the summary. RNN model with encoder decoder architecture was quite popular for text summarization as it is based on sequence to sequence model [13]. Lopyrev [14] proposed an encoder-decoder RNN with LSTM units and attention for producing headlines using the text from news articles. The model was found to be effective at concisely paraphrasing news articles. Sabahi[15] developed bidirectional RNN in order to make use of past and future and to generate balanced outputs as a result. The limitations of RNN were addressed by means of use of LSTM[16] and hybrid LSTM and Convolution Neural Network (CNN) [17] for summarization tasks. Currently transformer based models are gaining popularity in this text summarization. One such Transformer based model called BERT is discussed in detail in the next section.

## 3. BERT

BERT is a deep bidirectional representation trainer that uses both left and right context conditioning in all layers to train deep bidirectional representations from unlabeled text. As a result, the pre-trained BERT model may be fine-tuned with just one additional output layer to produce a cutting-edge model that can be used for tasks like Word Masking, Text Summarization, Text to Text Generation, Question Answering, and Language Interference [18]. BERT outperforms other Transformer architectures by eliminating the unidirectional requirement as it is pre-trained with a Masked Language Model (MLM). The masked model randomly masks out 10% to 15% of the words in the training data in the first step. This is done in order to learn to predict the masked words for the model, and the second step takes a sample sentence and a candidate sentence as an input, trying to predict whether the input sentence is properly followed by the candidate sentence [19]. The aim of the MLM, permits the representation word which incorporate the right and left context, allowing to pre-train transformer in both directions, unlike the pre-training of left-to-right language model. Additionally, BERT predicts the next sentence along-with the masked language model which enables pre-training of text-pair representations jointly.

BERT consists of two steps: Pre-training and Fine-tuning. Unlabeled data is used to train the model for various tasks, during the pre-training stage. The BERT model is fine-tuned on labeled data from downstream tasks after it has been initialized with the pre-trained parameters. Although they start with the same pre-trained features, each downstream task has its own fine-tuned model. BERT performs pre-training of unlabeled text by considering the conditions from left and right context and eventually provides bidirectional representations at all levels. As a consequence, the pre-trained BERT model may be fine-tuned with only one more output layer in order to deliver state-of-the-art models without requiring large task-specific architecture changes. BERT makes use of a fine-tuning-based technique for applying pre-trained language models; that is, a common architecture is trained for a somewhat generic job, and then it is fine-tuned on specific downstream tasks that are more or less comparable to the pre-training task.

The BERT architecture was chosen because it outperformed other NLP algorithms on sentence embedding in contrast to the word embedding .The sentence embedding retains the context of the sentence [18]. The word embedding techniques followed so far did not consider the fact that a given word can have entirely different meaning with respect to contexts used. The quality of generated results largely depends on appropriate understanding and differentiating among different senses of a word .This relationship among sentences was introduced by Devlin et al. [18] with BERT algorithm. The proposed work used for pre-training a language model was a transformer network for extracting the word [21]. Although BERT is based on the transformer concept, it has unique pre-training objectives. Approximately 10% to 15% words from the training set are chosen randomly. Later, they are getting masked off in first step. In the next step the attempts are made to predict the masked words. The model takes an input sentence and a candidate sentence, anticipating if the candidate sentence accurately follows the input sentence [19].Even with a large number of GPUs, this operation can take several days to complete. As a result, Google made two BERT models available to the public, one with 110 million parameters and the other with 340 million parameters [19].

The Transformers are based on the Encoder - Decoder architecture. The Encoder produces Embeddings and the Decoder receives these embeddings as input. The Encoding component is a stack of Encoders (Twelve in case of the Base model and Twenty-Four for the Large model). Each Encoder can be broken down into two components - the self attention layer: layer which helps the encoder to capture each and every word in the input sentence as it encodes a specific word. The self-attention layer outputs are then fed to the feed-forward neural network. Each location is equipped with the same feed-forward network. The Encoder looks for critical features in the input text and generates an embedding either for each word based on its relevance to other words in the sentence. The Decoder has the same components as the Encoder, but adds a "Attention layer" to aid the decoder in focusing on key sections of the input text. The Decoder takes the encoder's output, which is an embedding, and converts it back into text, which is the translated version of the input text.

## 4. BERT MODELS

BERT comes in two variants $BERT_{BASE}$ and $BERT_{LARGE}$. The $BERT_{BASE}$ model comprises 12 encoder layers piled on top of each other whilst the $BERT_{LARGE}$ model contains 24 layers. As the number of layers in $BERT_{LARGE}$ increases, so the number of parameters (weights) and number of attention heads also increase. $BERT_{BASE}$ comprises 12 attention heads with 110 million parameters. $BERT_{LARGE}$ on the other hand contains 16 attention heads and 340 million parameters. The $BERT_{BASE}$ model has 768 hidden layers as compared to $BERT_{LARGE}$ having 1024 hidden layers.

There are further two types of BERT models: *cased and uncased.* Cased models maintain the case of the Input sentence. The Uncased Model, on the other hand, converts the input sentence into lowercase before WordPeice Tokenization. The uncased model is not case sensitive i.e. difference between the words 'english' and 'English' is not considered. Based on the transformer based architectures BERT provides following variants:

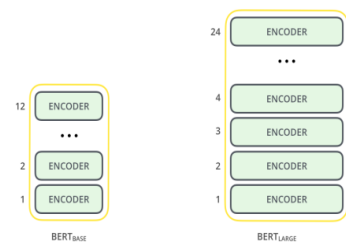

**Fig. 1.** BERT Variants

i. *BERTSUM* - A simple variant of BERT designed specifically for Text Summarization. It provides a score to each phrase based on how much value it adds to the total content. As a result, [s1,s2,s3] receives [score1, score2, score3]. The highest-scoring sentences are then gathered and rearranged to provide the article's overall summary.

ii. *BERT-LARGE-UNCASED* The larger version of BERT has 24 encoder layers. This model does not differentiate between English and english.

iii. *BERT-LARGE-CASED* The larger version of BERT has 24 encoder layers and it differentiates English and english.

iv. *BERT-BASE-UNCASED* - The smaller version of BERT that has 12 encoder layers and it does not differentiate between English and english.

v. *BERT-BASE-CASED* - The smaller version of BERT that has 12 encoder layers and differentiates between English and english.

vi. *DistilBERT* - A compact light variant of BERT mainly useful for reducing the cost of training an extractive summarizer and provide a summary while retaining the model's performance on less-resource devices. DistilBERT during the pre-training phase seeks to use knowledge distillation inorder to demonstrate that a BERT model compressed by 40% in size while still able to maintain maximum language comprehension skills with faster execution[22].

vii. *Pegasus-XSum* - Abstractive Summarization with Pretraining from Extracted Gap-sentences i.e PEGASUS is a model developed by Google which is specifically used for Abstractive Summarization. The model has been pre-trained on HugeNews and C4 Datasets. The C4 dataset is a corpus of 750GB of English-language text gathered from the public Common Crawl web scrape. It contains heuristics to extract only natural language.

## 5. BERT implementation and Evaluation metric for book summarization

The proposed work focuses on the standard BERT models (base and large, uncased), which are trained across a 3.3 billion word English corpus with a multi-task goal of masked language modeling and next-sentence prediction [20]. The proposed work follows Tenney et al. [20] and freezes the encoder weights to learn how the network reflects language as a result of pre-training. This stops the encoder from reorganizing its internal representations to fit the probing job better. BERT was used for chapter wise Extractive Summarization so that the Context of the Book isn't lost as it makes use of the surrounding text in order to establish context. The model can easily differentiate the same word in two different contexts; therefore it is able to provide two different vectors for the same word. Secondly, as the model has been pre-trained on a 3.3B word corpus, it can easily adapt to new references and words that are not even a part of its vocabulary. Hence the use of BERT for Chapter wise Summarization is recommended as the flow of the Book along with the context is maintained. The transformer model used for Chapter wise Extractive Summary was 'bert-extractive-summarizer' .The book used for demonstration was "Rich Dad Poor Dad" by Robert T Kiyosaki which contains 9 Chapters across 234 Pages having a total word count of 55,000 words.

The model takes a Portable Document Format (PDF) file as an input. Then text cleaning and stopword removal is performed, the cleaned text is then passed to the Transformer model for summarization. The model is pre-trained on MLM and NSP, and fine-tuned for Summarization. The use of HuggingFace Transformer pipeline is recommended, and various models can be explored for summarization using the same pipeline. By ensuring that the transformers and predictors are trained with the same samples, pipelines help prevent statistics from leaking from your test data into the trained model in cross-validation. After the text processing, the book is split into chapters with the help of PDF libraries. Extractive Summaries are generated for each chapter and then compiled together to form the Abstract of the Book. The Abstract is obtained by generating an Abstractive Summary of all the chapter wise summaries compiled together. The reason to use Abstractive Summarization instead of the Extractive Approach is to present the Abstract in a more Human Readable Format.

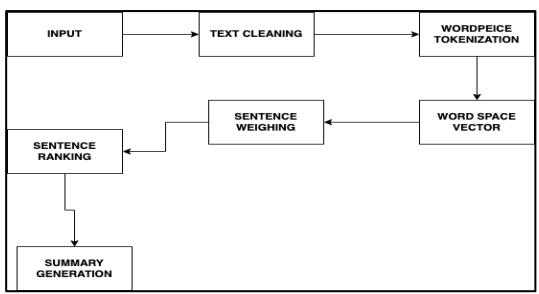

**Fig. 2.** Process Flow Model

Fig. 2 illustrates the Process flow of the model. The input is a PDF file which is sent for pre-processing where it is checked for stop words and punctuations, then word piece tokenization is performed.

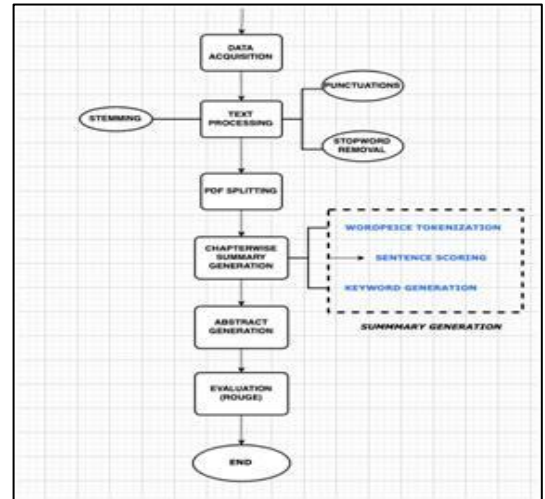

**Fig. 3.** Workflow of the Model

Word piece Tokenization is a sub word tokenization algorithm used by BERT, where each word is pre-tokenized by splitting on punctuations and whitespaces. This gives rise to a word space vector where each sub-unit is called a word piece. Later the sentences are weighed based on their importance in the summary; each sentence is assigned a rank based on its weight. The sentences are then selected to be included in the summary based on their ranks.

The Workflow diagram depicted in Fig.3. elaborates the process of summarization. First we acquire the Data, the data is then sent for pre-processing where stemming, stop word and punctuation removal is performed. The book is then split chapter wise, the split chapters are then sent to the model for extractive summary generation. The summaries of individual chapters are then sent to the BertSumAbs model for Abtractive Summarization. The generated summary is then evaluated under the ROUGE metric- by comparing it to a human-generated summary, the ROUGE -N score is calculated.

### 5.1.1. ROUGE Evaluation

ROUGE, or Recall-Oriented Understudy for Gisting Evaluation [23] compares a model-generated summary or phrase to a "gold-standard," often a human-generated, summary or sentence. ROUGE is the widely applied metric for evaluating the performance of any NLP model. There is no absolutely correct answer for summary generation. Relevant and important information to be included in the summary differs from individual to individual as each person perceives information differently.

Rouge-N evaluates by comparing the summary generated by human against the summary generated by the model using an *N-gram recall metric*. It simply compares the number of N-grams similar in the summary by human with that of summary by model. It measures the match between the handwritten summary and model produced summary. There are different variants of the ROUGE metric depending on number of overlapping words also called as grams that are measured for instance one word match called as ROUGE-1(uni-gram), ROUGE-2 for two words match (bi-gram), ROUGE-3 (trigram) for three words overlap, and ROUGE-L for Longest common sequence of words[24]. The proposed model uses ROUGE-N performance metric and got the following result. ROUGE needs reference (which is human generated summary) and predictions which is (model generated summary) as input and it produces precision, recall and F-score as output. The proposed model generated the following results for the given application.

**Table 2.** Comparison of ROUGE scores for book abstraction for proposed model and human generated abstract

|  | ROUGE-1 | ROUGE-2 | ROUGE-L |
|---|---|---|---|
| *Precision* | 0.543 | 0.564 | 0.58 |
| *Recall* | 0.529 | 0.478 | 0.467 |
| *F-Score* | 0.535 | 0.517 | 0.527 |

The results tabulated in table 1 clearly indicate that almost 50% system generated abstract overlap with the human generated abstract. The model needs to be improvised by providing a rich vocabulary in order to improve the text generation to cope with human cognitive level. The results are found to be promising but the results are not compared as no prior work found for this application.

## 6. Conclusion

This paper is an attempt to apply pre-trained BERT model for book summarization. For every chapter in our corpus we compare the model generated summary with a human-generated summary. The sequential summarization - extractive of each individual chapter and abstractive for the compiled summaries of all the chapters, proves to be effective as compared to using only extractive or only abstractive in many ways as in extractive Summarization we pick the sentences from the document as it is; thus leading to retain the essential vocabulary related to the book and the wordings of the book are not altered. Similarly, using the abstractive approach for final abstract generation is simply an attempt to generate a concise summary using the least number of words possible and doing so also gives a human touch to the summary which is not possible with the extractive approach.

## 7. References and Footnotes

**Author contributions**

**Siddhant Porwal:** Data generation, Methodology, Software, Evaluation, Writing-Original draft preparation
**Laxmi Bewoor:** Conceptualization, Methodology, Revising Original draft, Software, Evaluation and Validation.,
**Vivek Deshpande:** Investigation, Reviewing, Suggestions

**Conflicts of interest**

The authors declare no conflicts of interest.

## References

[1]. N. Andhale and L. A. Bewoor, "An overview of text summarization techniques," 2017, doi: 10.1109/ICCUBEA.2016.7860024.

[2]. N. S. Shirwandkar and S. Kulkarni, "Extractive Text Summarization Using Deep Learning," Proc. - 2018 4th Int. Conf. Comput. Commun. Control Autom. ICCUBEA 2018, 2018, doi: 10.1109/ICCUBEA.2018.8697465.

[3]. S. Singhal, "Abstractive Text Summarization," J. Xidian Univ., vol. 14, no. 6, pp. 1–11, 2020, doi: 10.37896/jxu14.6/094..

[4]. A. Vaswani et al., "Attention is all you need," Adv. Neural Inf. Process. Syst., vol. 2017-December, no. Nips, pp. 5999–6009, 2017.

[5]. S. Singh and A. Mahmood, "The NLP Cookbook: Modern Recipes for Transformer Based Deep Learning Architectures," IEEE Access, vol. 9, pp. 68675–68702, 2021, doi: 10.1109/ACCESS.2021.3077350.

[6]. J. Devlin, M. W. Chang, K. Lee, and K. Toutanova, "BERT: Pre-training of deep bidirectional transformers for language understanding," *NAACL HLT 2019 - 2019 Conf. North Am. Chapter Assoc. Comput. Linguist. Hum. Lang. Technol. - Proc. Conf.*, vol. 1, no. Mlm, pp. 4171–4186, 2019.

[7]. R. Mihalcea and H. Ceylan, "Explorations in automatic book summarization," *EMNLP-CoNLL 2007 - Proc. 2007 Jt. Conf. Empir. Methods Nat. Lang. Process. Comput. Nat. Lang. Learn.*, no. June, pp. 380–389, 2007.

[8]. J. Wu *et al.*, "Recursively Summarizing Books with Human Feedback," 2021, Available: http://arxiv.org/abs/2109.10862.

[9]. H. P. Luhn, "The Automatic Creation of Literature Abstracts," *IBM J. Res. Dev.*, vol. 2, no. 2, pp. 159–165, 2010, doi: 10.1147/rd.22.0159.

[10]. A. Kumar, A. Sharma, S. Sharma, and S. Kashyap, "Performance analysis of keyword extraction algorithms assessing extractive text summarization," *2017 Int. Conf. Comput. Commun. Electron. COMPTELIX 2017*, pp. 408–414, 2017, doi: 10.1109/ COMPTELIX.2017.8004004.

[11]. R. Bhargava and Y. Sharma, "Deep Extractive Text Summarization," *Procedia Comput. Sci.*, vol. 167, no. 2019, pp. 138–146, 2020, doi: 10.1016/j.procs.2020.03.191.

[12]. S. Syed, Abstractive Summarization of Social Media Posts: A case Study using Deep Learning, Master's thesis, Bauhaus University, Weimar, Germany, 2017.

[13]. I. Sutskever, O. Vinyals, and Q. V. Le, "Sequence to sequence learning with neural networks," Adv. Neural Inf. Process. Syst., vol. 4, no. January, pp. 3104–3112, 2014.

[14]. K. Lopyrev, Generating news headlines with recurrent neural networks, p. 9, 2015, https://arxiv.org/abs/1512.01712.

[15]. K. Al-Sabahi, Z. Zuping, and Y. Kang, Bidirectional Attentional Encoder-Decoder Model and Bidirectional Beam Search for Abstractive Summarization, Cornell

University, Ithaca, NY, USA, 2018, http://arxiv.org/abs/1809.06662.

[16]. Tomer, M., Kumar, M. Improving Text Summarization using Ensembled Approach based on Fuzzy with LSTM. *Arab J Sci Eng* 45, 10743–10754 (2020). https://doi.org/10.1007/s13369-020-04827-6

[17]. Song, S., Huang, H. & Ruan, T. Abstractive text summarization using LSTM-CNN based deep learning. *Multimed Tools Appl* 78, 857–875 (2019). https://doi.org/10.1007/s11042-018-5749-3

[18]. D. Miller, "Leveraging BERT for Extractive Text Summarization on Lectures," 2019, Available: http://arxiv.org/abs/1906.04165.

[19]. J. Devlin, M. W. Chang, K. Lee, and K. Toutanova, "BERT: Pre-training of deep bidirectional transformers for language understanding," *NAACL HLT 2019 - 2019 Conf. North Am. Chapter Assoc. Comput. Linguist. Hum. Lang. Technol. - Proc. Conf.*, vol. 1, no. Mlm, pp. 4171–4186, 2019.

[20]. I. Tenney, D. Das, and E. Pavlick, "BERT rediscovers the classical NLP pipeline," *ACL 2019 - 57th Annu. Meet. Assoc. Comput. Linguist. Proc. Conf.*, pp. 4593–4601, 2020, doi: 10.18653/v1/p19-1452.

[21]. S. Zoupanos, S. Kolovos, A. Kanavos, O. Papadimitriou, and M. Maragoudakis, "Efficient comparison of sentence embeddings," 2022. Available: http://arxiv.org/abs/2204.00820.

[22]. V. Sanh, L. Debut, J. Chaumond, and T. Wolf, "DistilBERT, a distilled version of BERT: smaller, faster, cheaper and lighter," pp. 2–6, 2019, [Online]. Available: http://arxiv.org/abs/1910.01108.

[23]. Lin, Chin-Yew. ROUGE: A Package for Automatic Evaluation of Summaries. In Proceedings of the Workshop on Text Summarization Branches Out (WAS 2004), Barcelona, Spain, July 25 - 26, 2004.

[24]. D. Suleiman and A. Awajan, "Deep Learning Based Abstractive Text Summarization: Approaches, Datasets, Evaluation Measures, and Challenges," *Math. Probl. Eng.*, vol. 2020, 2020, doi: 10.1155/2020/9365340.